# Highly Accurate CNN Inference Using Approximate Activation Functions over Homomorphic Encryption

Takumi Ishiyama
Dept. of Computer Science and
Communications Engineering
Waseda University
Tokyo, Japan
takumi@yama.info.waseda.ac.jp

Takuya Suzuki
Dept. of Computer Science and
Communications Engineering
Waseda University
Tokyo, Japan
t-suzuki@yama.info.waseda.ac.jp

Hayato Yamana
Faculty of Science and Engineering
Waseda University
Tokyo, Japan
yamana@yama.info.waseda.ac.jp

*Abstract*—In the big data era, cloud-based machine learning as a service (MLaaS) has attracted considerable attention. However, when handling sensitive data, such as financial and medical data, a privacy issue emerges, because the cloud server can access clients' raw data. A common method of handling sensitive data in the cloud uses homomorphic encryption, which allows computation over encrypted data without decryption. Previous research adopted a low-degree polynomial mapping function, such as the square function, for data classification. However, this technique results in low classification accuracy. This study seeks to improve the classification accuracy for inference processing in a convolutional neural network (CNN) while using homomorphic encryption. We apply various orders of the polynomial approximations of Google's Swish and ReLU activation functions. We also adopt batch normalization to normalize the inputs for the approximated activation functions to fit the input range to minimize the error. We implemented CNN inference labeling over homomorphic encryption using the Microsoft's Simple Encrypted Arithmetic Library (SEAL) for the Cheon–Kim–Kim–Song (CKKS) scheme. The experimental evaluations confirmed classification accuracies of 99.29% and 81.06% for MNIST and CIFAR-10, respectively, which entails 0.11% and 4.69% improvements, respectively, over previous methods.

## I. INTRODUCTION

Cloud-based machine learning as a service (MLaaS) is an ideal solution for clients that lack high-computing facilities to run deep-learning tasks. However, when handling sensitive data, such as financial information and medical images, an important privacy issue emerges, because the cloud server that provides MLaaS can access clients' raw data. Thus, a privacy-preserving machine learning (PPML) capability, which conducts training and inference processing using a machine-learning model while protecting privacy, is being actively pursued.

There are three typical approaches to PPML research. First, multi-party computation (MPC) uses garbled circuits [1] and secret sharing [2] to communicate and evaluate functions while concealing information from outside parties. Second, homomorphic encryption (HE) [3, 4] allows computation on encrypted data without decryption. Third, a trusted execution environment [6] executes codes while protecting private data in a cryptographically-protected hardware enclave, such as with Intel SGX [5]. In this research, we focus on the use of HE.

A HE scheme adopting bit-wise encryption performs arbitrary operations; however, it has a drawback of long execution time. Fortunately, a method adopts a HE scheme to encrypt integers or complex numbers to shorten execution time. However, it cannot handle functions that apply comparisons, divisions, and exponential operations. Therefore, rectified linear units (ReLU), Sigmoid functions, and max-pooling operations, commonly used in deep learning, cannot be computed. Thus, recent works on PPML over HE have focused on using the approximate functions for polynomials [7, 8, 9, 10, 13]. Then they adopted a fully HE (FHE) over-the-torus (TFHE) scheme that allows both bit-wise encryption and evaluation of arbitrary Boolean circuits comprising binary gates with binary neural networks to accelerate inferencing [11, 12].

In 2018, Badawi et al. [14] accelerated the inference of a convolutional neural network (CNN) using a graphical processing unit for the first time. They achieved a high classification accuracy (99%) on the Modified National Institute of Standards and Technology (MNIST) dataset [15]. However, they achieved low classification accuracy (77.55%) on the Canadian Institute For Advanced Research (CIFAR)-10 dataset [16]. This is caused by the use of a square function as the activation function.

In this paper, based on the results of Badawi et al. [14], we propose a new method to improve the accuracy of CNNs over HE. Note that the accuracy over HE is not enough in the previous studies; thus, improving the accuracy is challenging. We adopt a polynomial approximation of Google's Swish activation function and the ReLU, and we apply batch normalization (BN). Swish is a smooth, non-monotonic function that consistently matches or outperforms the ReLU activation function on deep networks when applied to image classification and machine translation domains. Besides, by adopting BN, we normalize the inputs for the approximated activation function to fit the input range to minimize the error.

The rest of this paper is organized as follows. Section 2 describes an overview of HE, followed by related work on the inference of CNN over HE in Section 3. Section 4 proposes our method. The results of an experimental evaluation follow in Section 5. Finally, a conclusion is provided in Section 6.

## II. Homomorphic Encryption (HE)

Homomorphic encryption (HE) enables both homomorphic addition and homomorphic multiplication without decrypting the ciphertext. With HE, the difficulty of deciphering the original data is secured by adding random noise to the ciphertext. This noise increases every time the homomorphic operation is performed, and, if the amount of the noise exceeds a threshold, ciphertexts cannot be correctly decrypted.

Presently, fully homomorphic encryption (FHE) handles bootstrapping which resets the noise but requires a large time-complexity and space-complexity. Therefore, when applying HE to machine-learning applications such as CNNs, it is appropriate to use leveled HE (LHE), which pre-determines the number of multiplications without bootstrapping. Additionally, the larger the level, $l$, (the maximum number of homomorphic multiplications that can be applied to ciphertext), the slower the execution speed, and the larger the memory usage. Thus, it is necessary to reduce the consumption level by minimizing the multiplications.

Typical LHE schemes include Brakerski–Gentry–Vaikuntanathan (BGV) [21], Brakerski/Fan-Vercauteren (BFV) [22, 23, 24], and Cheon–Kim–Kim–Song (CKKS) schemes [4, 25]. BGV and BFV schemes are suitable for representing Integers. In this study, we adopt the CKKS method, which can approximately represent real numbers, because it is indispensable for machine-learning applications.

The BGV, BFV, and CKKS schemes support the packing mechanism to pack multiple data into one ciphertext [26]. It can perform single-instruction multiple-data (SIMD) calculations and improve throughput, which is why we apply it in this study.

## III. Related Work

In this section, recent studies that used polynomial approximation activation functions for inferencing CNNs over HE are described. Note that the related works and ours both assume that the cloud server already has a trained model in advance. Because the cloud server does not decrypt the encrypted data, client privacy is protected. Fig. 1 outlines the common inference process, which applies the following steps:

1. A client generates a set of public and secret keys and encrypts data with the public key.
2. The client sends both the encrypted data and the public key to the cloud server.
3. The cloud server performs inference over HE using the pre-trained model.
4. The cloud server calculates the encrypted inference result.
5. The cloud server sends the encrypted result to the client.
6. The client decrypts the encrypted result using the private key to obtain the inference result.

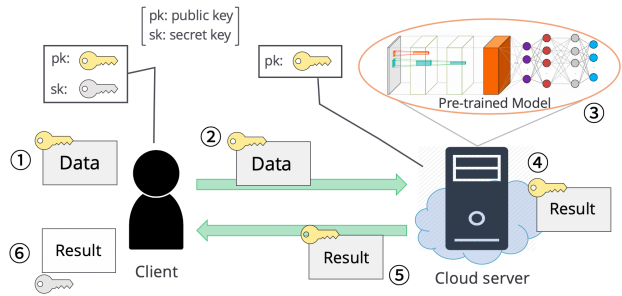

Fig. 1. Outline of inference on HE

TABLE I. Recent work on inferencing CNNs using HE

| | Scheme | Activation | Pooling | BN | Accuracy [%] | |
|---|---|---|---|---|---|---|
| | | | | | *MNIST* | *CIFAR-10* |
| Dowlin et al. [7] | YASHE | Square | Scaled mean | - | 98.95 | - |
| Chabanne et al. [8] | BGV | ReLU (4-degrees) | Average | ✓ | 99.30 | - |
| Jiang et al. [9] | CKKS | Square | - | - | 98.10 | - |
| Hesamifard et al. [10] | BGV | ReLU (3-degrees) | Scaled mean | - | 99.25 | - |
| Chow et al. [13] | BFV | ReLU/Swish (2-degrees) | Scaled mean/Average | ✓ | 99.10 | 75.99 |
| Badawi et al. [14] | BFV | Square | Average | - | 99 | 77.55 |

Table I summarizes recent research on inferencing CNNs over HE. The problem of the existing studies is that they achieved high classification accuracies for the MNIST dataset, but they achieved low accuracies for the CIFAR-10 dataset. In existing deep-learning studies on plaintext applications, the classification accuracy of the CIFAR-10 dataset achieved generally greater than 90%. Thus, looking at Table I, 75.99, and 77.55% are low rates. We assume that the reason is that the activation functions are low degrees of polynomial approximations. Thus, a new method that obtains enough classification accuracy is needed.

## IV. Proposed Method

To improve the classification accuracy of CNN inference over HE, we apply the activation functions that approximate Swish [17] and ReLU with a polynomial. Besides, we apply batch normalization (BN) to normalize the inputs for the approximate functions to fit the input range to minimize the error. As for the HE related optimizations, two optimizations are applied to reduce the consumption level of LHE to speed up.

### A. Polynomial approximation of activation function

We apply Swish [17] in addition to ReLU to confirm the effectiveness when their polynomial approximations with various degrees are adopted. Swish is a versatile activation function found in studies that replaced the ReLU, where $f(x) = \max(0, x)$. Swish is a Sigmoid function multiplied by $x$ and is expressed as $f(x) = \frac{x}{1+e^{-x}}$.

By applying BN (described in the next sub-section) just before an activation function, the input value of the activation function follows a Gaussian distribution. Thus, we approximate the Swish and the ReLU in the x-axis ranges of $[-4, 4]$ or $[-6, 6]$, using a second or fourth-order polynomial, because an error with the approximated function can be minimized in the definition range around 0. We use the *leastsq*[1] function provided by the SciPy library [30] of Python for polynomial approximation. In order to suppress the consumption level, it is necessary to approximate with a polynomial of low order. We approximated Swish at various orders as shown in Fig.2. Fig. 2 shows that the fourth-order polynomial approximates Swish exactly in the specified definition range. Therefore, we chose to use a fourth-order polynomial for approximation. However, we also use a second-order polynomial approximation activation function for comparison. We also approximate ReLU at various orders shown in Fig.3. Table II shows the results of the polynomial approximation of Swish and ReLU. Fig. 4 shows a comparison between Swish and its polynomial approximation, and Fig. 5 shows a comparison between ReLU and its polynomial approximation in the x-axis range of $[-4, 4]$ and $[-6, 6]$.

TABLE II. POLYNOMIAL APPROXIMATION RESULT

| Activation function | degree | x-axis range to be fitted | polynomial approximated |
|---|---|---|---|
| Swish | 4 | $[-4, 4]$ | $0.03347 + 0.5x + 0.19566x^2 - 0.005075x^4$ |
| | | $[-6, 6]$ | $0.1198 + 0.5x + 0.1473x^2 - 0.002012x^4$ |
| | 2 | $[-4, 4]$ | $0.12592 + 0.5x + 0.145276x^2$ |
| | | $[-6, 6]$ | $0.0851505 + 0.5x + 0.344125x^2$ |
| ReLU | 4 | $[-4, 4]$ | $0.234606 + 0.5x + 0.204875x^2 - 0.0063896x^4$ |
| | | $[-6, 6]$ | $0.119782 + 0.5x + 0.147298x^2 - 0.002015x^4$ |
| | 2 | $[-4, 4]$ | $0.375373 + 0.5x + 0.117071x^2$ |
| | | $[-6, 6]$ | $0.563059 + 0.5x + 0.078047x^2$ |

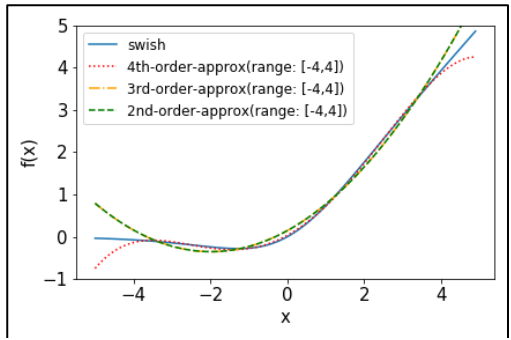

Fig. 2. Comparison of Swish and its polynomial approximation with various orders

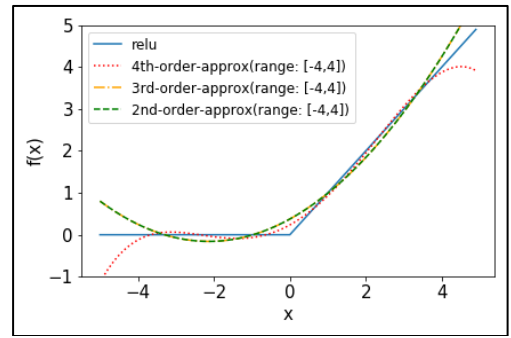

Fig. 3. Comparison of ReLU and its polynomial approximation with various orders

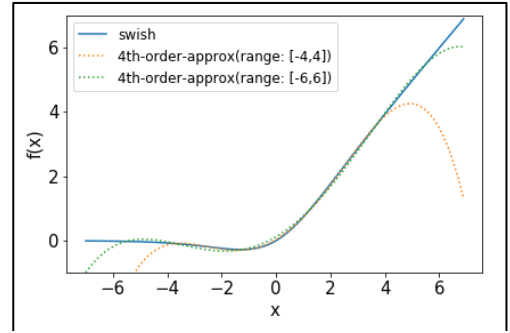

Fig. 4. Comparison of Swish and Swish polynomial approximation

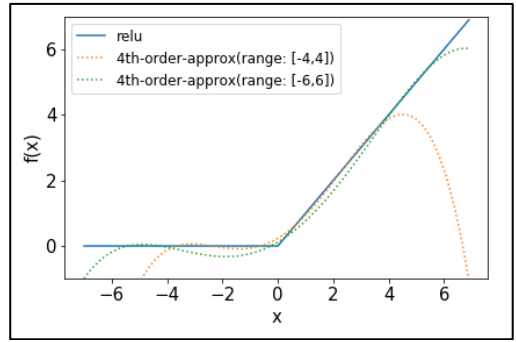

Fig. 5. Comparison of ReLU and ReLU polynomial approximation

## B. Batch normalization

BN [18] is a method generally used in deep learning applications. Using the average and variance of the layer output of each element (channel or unit) in a mini-batch prevents the distribution of variables inside the network (internal covariate shift) from changing significantly. This standardization both improves classification accuracy and accelerates learning convergence. Algorithm 1 shows the batch normalizing transform steps.

In Algorithm 1, the $i$-th output $y_i$ in the mini-batch is represented by formula (1). In the inference phase, it is not necessary to calculate the mean and variance of the mini-batches ($\mu_B$ and $\sigma_B^2$) because the estimated values of the population mean ($\mu_{pop}$) and the population variance ($\sigma^2_{pop}$) are calculated during training. In other words, $\mu_{pop}$, $\sqrt{\sigma^2_{pop} + \epsilon}$, $\gamma$, and $\beta$ can be obtained from the trained model; thus, the BN process in the inference is expressed by formula (2), where the weight parameter $W_{BN}$ is $\frac{\gamma}{\sqrt{\sigma^2_{pop}+\epsilon}}$ and the bias parameter $B_{BN}$ is $\beta - \frac{\gamma\mu_{pop}}{\sqrt{\sigma^2_{pop}+\epsilon}}$, both can be pre-computed. Formula (2) shows that processing the BN over HE consumes one level.

---

[1] scipy.optimize.leastsq, SciPy.org,
https://docs.scipy.org/doc/scipy/reference/generated/scipy.optimize.leastsq.html

$$y_i = \frac{\gamma}{\sqrt{\sigma_B^2 + \epsilon}}(x_i - \mu_B) + \beta \quad (1)$$

$$y = W_{BN}x + B_{BN}. \quad (2)$$

**Algorithm 1 Batch Normalizing Transform (Revised from [18])**

**Input:** Mini-batch $B = \{x_{1...m}\}$, trainable parameter $\gamma, \beta$
**Output:** $y = \{y_{1...m}\}$
1. $\mu_B \leftarrow \frac{1}{m}\sum_{i=1}^{m} x_i$      // mini-batch mean
2. $\sigma_B^2 \leftarrow \frac{1}{m}\sum_{i=1}^{m}(x_i - \mu_B)^2$   // mini-batch variance
3. $for\ i \in \{1, 2, 3, \dots, m\}\ do$
4.     $\hat{x}_i \leftarrow \frac{x_i - \mu_B}{\sqrt{\sigma_B^2 + \epsilon}}$   // normalize ($\epsilon$ is a constant value)
5.     $y_i \leftarrow \gamma\hat{x}_i + \beta$ // scale and shift
6. $end\ for$
7. $return\ y$

*C. Optimizations to reduce consumption level*

The fourth-order polynomial approximation consumes three levels, besides one more level consumed by BN. Thus, four levels are consumed in total. Our naive method without optimization consumes three more levels for each evaluation of the activation function than the previous method using the square function as an activation function without BN. To reduce the consumption level (i.e., the number of multiplications), we apply the following two optimizations. They are commonly used in deep-learning optimization and have been used in compilers for deep learning using HE [28, 29].

The first optimization fuses the convolutional layer with BN. In the convolutional layer, we consider the process of convolving a certain $s \times s \times c$ filter and the input, $x$, corresponding to the filter to obtain the output, $y$. The weights and biases of the convolutional layer obtained from the trained model are $W_{Conv}$ ($s \times s \times c$) (matrix) and $B_{Conv}$ (scalar), respectively. The constant parameters, $W_{BN}$ and $B_{BN}$, of BN are shown in formula (2), and the inference of convolutional layer and BN can be collectively expressed using formula (3), where $x^{i,j,k}$ shows an element at the $i$-th row, $j$-th column, and $k$-th channel of inputs. $W_{Conv}^{i,j,k}$ shows an element at the $i$-th row, $j$-th column, and $k$-th channel of filter weights. $W_{Conv}^{i,j,k}W_{BN}$ and $B_{Conv}W_{BN} + B_{BN}$ are calculated in advance. Therefore, we can apply BN without increasing the level required.

$$y = W_{BN}\left(\sum_{i=0}^{s}\sum_{j=0}^{s}\sum_{k=0}^{c} W_{Conv}^{i,j,k} x^{i,j,k} + B_{Conv}\right) + B_{BN}$$

$$= \left(\sum_{i=0}^{s}\sum_{j=0}^{s}\sum_{k=0}^{c} W_{Conv}^{i,j,k} W_{BN} x^{i,j,k}\right) + (B_{Conv}W_{BN} + B_{BN}) \quad (3)$$

The second optimization is the manipulation of coefficients in the Swish polynomial approximation, which is expressed as $f(x) = ax^4 + bx^2 + cx + d$. Here, we divide both sides by the coefficient, $a$, of the highest order term, and $f'(x) = \frac{f(x)}{a}$, $b' = \frac{b}{a}$, $c' = \frac{c}{a}$, $d' = \frac{d}{a}$. Thus, we have $f'(x) = x^4 + b'x^2 + c'x + d'$. The consumption level of function $f'$ is two. Because $f'$ is the original function, $f$, multiplied by $\frac{1}{a}$, the weight parameters of the convolutional, pooling, and fully connected layers immediately after applying the activation function are multiplied by $a$ in advance.

By applying the above two optimizations, the level required for evaluating the Swish polynomial approximation and BN is reduced by two, which results in only one level increase from the evaluation of the square function.

V. EXPERIMENTAL EVALUATION

We conducted experiments to evaluate our proposed method on both the MNIST and the CIFAR-10 datasets. We compared the proposed methods with the baseline method that uses the square function as the activation function.

*A. Dataset*

We used the MNIST dataset and the CIFAR-10 dataset.

The MNIST dataset consists of 70,000 handwritten images (60,000 training data and 10,000 test data), each of which is a $28 \times 28$ pixel grayscale image. Each image is labeled with one of 10 classes from 0 to 9. The CIFAR-10 dataset consists of 60,000 images (50,000 training data and 10,000 test data), each of which is a $3 \times 32 \times 32$ pixel RGB image. Each image is labeled with one of 10 classes.

*B. Network architecture*

The CNN used in our experiment followed the same architecture used by Badawi et al. [14]. Table III shows the architecture for MNIST, and Table IV shows the architecture for CIFAR-10. The difference from [14] is that BN is added immediately after each convolutional layer in the proposed method, shown with closed parentheses in Tables III and IV.

TABLE III. CNN ARCHITECTURE FOR MNIST DATASET (BATCH NORMALIZATION IS ADOPTED ONLY IN THE PROPOSED METHOD)

| Layer | Parameters | Output size |
|---|---|---|
| Convolution | 5 filters of size 5x5, (2, 2) stride, no padding | 12x12x5 |
| (Batch Normalization) | - | 12x12x5 |
| Activation | - | 12x12x5 |
| Convolution | 50 filters of size 5x5, (2, 2) stride, no padding | 4x4x50 |
| (Batch Normalization) | - | 4x4x50 |
| Activation | - | 4x4x50 |
| Fully Connected | 10 units | 1x1x10 |

TABLE IV. CNN ARCHITECTURE FOR CIFAR-10 DATASET (BATCH NORMALIZATION IS ADOPTED ONLY IN THE PROPOSED METHOD)

| Layer | Parameters | Output size |
|---|---|---|
| Convolution | 32 filters of size 3x3, (1, 1) stride, (1, 1) padding | 32x32x32 |
| (Batch Normalization) | - | 32x32x32 |
| Activation | - | 32x32x32 |
| Average Pooling | pool size 2x2, (2, 2) stride | 16x16x32 |
| Convolution | 64 filters of size 3x3, (1, 1) stride, (1, 1) padding | 16x16x64 |
| (Batch Normalization) | - | 16x16x64 |
| Activation | - | 16x16x64 |
| Average Pooling | pool size 2x2, (2, 2) stride | 8x8x64 |
| Convolution | 128 filters of size 3x3 (1, 1) stride, (1, 1) padding | 8x8x128 |
| (Batch Normalization) | - | 8x8x128 |
| Activation | - | 8x8x128 |
| Average Pooling | pool size 2x2, (2, 2) stride | 4x4x128 |
| Fully Connected | 256 units | 1x1x256 |
| Fully Connected | 10 units | 1x1x10 |

### C. Accuracy in plaintext

Because the CKKS scheme approximates complex numbers, a small error occurs from the original data. It is expected that the classification accuracy over ciphertext will be lower than that of plaintext. Therefore, to confirm the differences, we performed inferencing on the plaintext test data.

We trained the model based on the network architecture shown in Table III and Table IV using the Keras library [31] with various activation functions (i.e., Swish, ReLU, Square, approximated Swish, and approximated ReLU). For MNIST, we set the parameters for training such that batch size as 128, epoch size as 150, and the optimizer as default parameter Adadelta[2]. For CIFAR-10, we set the parameters for training such that batch size as 256, epoch size as 150, and the optimizer as default parameter Adam[3]. We also applied data augmentation in training both datasets.

Table V shows the accuracy of MNIST and CIFAR-10. Table V shows that adapting BN to the Swish function improves the classification accuracy by 4.12% in CIFAR-10, whereas applying BN to its polynomial approximated activation function improves 5.17 to 6.22% of the accuracy. The same improvements are confirmed for the ReLU function; 0.16% improvement for the ReLU but 2.18 to 7.27% for its polynomial approximated activation function. Thus, we confirm that combining a polynomial activation function with BN is effective in improving the classification accuracy.

TABLE V. PLAINTEXT EXPERIMENT RESULT

| Activation function | Approx. degree | Approx. range | BN | Accuracy [%] MINST | Accuracy [%] CIFAR-10 |
|---|---|---|---|---|---|
| Swish | - | - | - | 98.86 | 77.88 |
| Swish | - | - | ✔ | 99.25 | 82.00 |
| ReLU | - | - | - | 98.82 | 83.53 |
| ReLU | - | - | ✔ | 99.05 | 83.69 |
| Square | - | - | - | 99.18 | 76.37 |
| Square | - | - | ✔ | 99.18 | 77.46 |
| Approx. Swish | 4 | $x \in [-4, 4]$ | - | 98.83 | 74.85 |
| Approx. Swish | 4 | $x \in [-4, 4]$ | ✔ | 99.16 | 80.09 |
| Approx. Swish | 4 | $x \in [-6, 6]$ | - | 98.79 | 75.30 |
| Approx. Swish | 4 | $x \in [-6, 6]$ | ✔ | 99.21 | 80.47 |
| Approx. Swish | 2 | $x \in [-4, 4]$ | - | 98.80 | 72.89 |
| Approx. Swish | 2 | $x \in [-4, 4]$ | ✔ | 99.28 | 79.11 |
| Approx. Swish | 2 | $x \in [-6, 6]$ | - | 98.46 | 70.58 |
| Approx. Swish | 2 | $x \in [-6, 6]$ | ✔ | 99.05 | 76.75 |
| Approx. ReLU | 4 | $x \in [-4, 4]$ | - | 98.90 | 74.43 |
| Approx. ReLU | 4 | $x \in [-4, 4]$ | ✔ | 99.11 | 79.38 |
| Approx. ReLU | 4 | $x \in [-6, 6]$ | - | 98.94 | 73.70 |
| Approx. ReLU | 4 | $x \in [-6, 6]$ | ✔ | 99.25 | 81.02 |
| Approx. ReLU | 2 | $x \in [-4, 4]$ | - | 98.51 | 71.80 |
| Approx. ReLU | 2 | $x \in [-4, 4]$ | ✔ | 99.17 | 79.07 |
| Approx. ReLU | 2 | $x \in [-6, 6]$ | - | 98.41 | 72.54 |
| Approx. ReLU | 2 | $x \in [-6, 6]$ | ✔ | 99.15 | 74.72 |

### D. Evaluation methods

We implemented the CNN inference program over HE using Microsoft's Simple Encrypted Arithmetic Library (SEAL) library [27], which implements the CKKS scheme.

When adopting CKKS, we rounded $x \in [0.0, 10^{-7})$ up to $10^{-7}$, and $x \in (-10^{-7}, 0.0)$ up to $-10^{-7}$ because (1) we cannot obtain a ciphertext after the multiplication of a plaintext whose value is 0 and a ciphertext; (2) we cannot use a number with a too-small absolute value for model parameters in the CKKS scheme because the CKKS scheme handles fixed-point numbers.

Similar to Badawi et al. [14], pixels of the same position in multiple images were encrypted in a single ciphertext using packing, and the inference was performed simultaneously in a SIMD manner. The activation function was evaluated by the square function or the polynomial approximation of Swish and ReLU, as shown in Table II. We compared the classification accuracy, execution time, and memory usage. Note that the execution time was averaged by the three continuous executions after the first execution. The baseline method uses a square function as the activation function, while our proposed method uses approximated Swish and ReLU as the activation function and apply the two optimizations.

---

[2] keras.optimizers.Adadelta, Keras, https://keras.io/api/optimizers/adadelta/

[3] keras.optimizes.Adam, Keras, https://keras.io/api/optimizers/adam/

We used four Xeon E7-8880 v3(2.30 GHz) with 72 cores in total and 3TB main memory for inferencing over HE. We used the parameters shown in Tables VI and VII for MNIST and CIFAR-10, respectively.

TABLE VI. SEAL PARAMETER SETTINGS for MNIST

|  | Activation function | Approx. degree | N | scale factor [bit] | log Q [bit] | level |
|---|---|---|---|---|---|---|
| Baseline | Square | - | $2^{14}$ | 30 | 200 | 5 |
| Proposed | Approx. Swish/ReLU | 4 | $2^{14}$ | 30 | 260 | 7 |
|  |  | 2 | $2^{14}$ | 30 | 200 | 5 |

TABLE VII. SEAL PARAMETER SETTINGS FOR CIFAR-10

|  | Activation function | Approx. degree | N | scale factor [bit] | log Q [bit] | level |
|---|---|---|---|---|---|---|
| Baseline | Square | - | $2^{14}$ | 30 | 290 | 8 |
| Proposed | Approx. Swish/ReLU | 4 | $2^{14}$ | 30 | 380 | 11 |
|  |  | 2 | $2^{14}$ | 30 | 290 | 8 |

*E. Evaluation Results on MNIST dataset*

Table VIII shows the result of the MNIST dataset. We measured using 16 threads with OpenMP. As a result of the evaluation, we achieved the best classification accuracy of 99.29%, which advances 0.11% compared with the baseline method, i.e., using the square function as the activation function. Thanks to packing, we classified 8,192 images with one inference. Thus, the amortized time (i.e., inference execution time per image) in the proposed method achieved the best accuracy (approximating Swish by a 2-degree polynomial in the range $[-4, 4]$) was 2.58 ms.

TABLE VIII. RESULTS ON MNIST DATASET (16 THREADS WERE USED.)

|  | Act. function | Approx. degree | Approx. range | level | Mem. usage [GB] | Acc. [%] | Infer. exec. time [s] |
|---|---|---|---|---|---|---|---|
| Baseline | Square | - | - | 5 | 17.5 | 99.18 | 20.42 |
| Proposed | Approx. Swish | 4 | $[-4, 4]$ | 7 | 21.3 | 99.16 | 32.66 |
|  |  |  | $[-6, 6]$ | 7 | 21.3 | 99.22 | 32.93 |
|  |  | 2 | $[-4, 4]$ | 5 | 16.2 | **99.29** | 21.15 |
|  |  |  | $[-6, 6]$ | 5 | 16.2 | 99.05 | 21.43 |
|  | Approx. ReLU | 4 | $[-4, 4]$ | 7 | 21.3 | 99.13 | 33.39 |
|  |  |  | $[-6, 6]$ | 7 | 21.3 | 99.24 | 33.21 |
|  |  | 2 | $[-4, 4]$ | 5 | 16.2 | 99.17 | 21.15 |
|  |  |  | $[-6, 6]$ | 5 | 16.2 | 99.15 | 21.22 |

*F. Evaluation Results on CIFAR-10 dataset*

Table IX shows the evaluation results on the CIFAR-10 dataset. We measured using 72 threads with OpenMP because of long execution times when using smaller numbers of threads. We achieved the best classification accuracy of 81.06% for the CIFAR-10 dataset, which improved accuracy by 4.69% compared with the baseline method. Similar to the MNIST experiment, the amortized time in the proposed method achieved the best accuracy (approximating ReLU by a 4-degree polynomial in the range $[-6, 6]$) was 190 ms.

TABLE IX. RESULTS ON CIFAR-10 DATASET (72 THREADS WERE USED.)

|  | Act. function | Approx. degree | Approx. range | level | Mem. usage [TB] | Acc. [%] | Infer. exec. time [s] |
|---|---|---|---|---|---|---|---|
| Baseline | Square | - | - | 8 | 1.16 | 76.37 | 1038.0 |
| Proposed | Approx. Swish | 4 | $[-4, 4]$ | 11 | 1.41 | 80.05 | 1554.5 |
|  |  |  | $[-6, 6]$ | 11 | 1.41 | 80.48 | 1484.3 |
|  |  | 2 | $[-4, 4]$ | 8 | 1.15 | 79.12 | 1039.1 |
|  |  |  | $[-6, 6]$ | 8 | 1.15 | 76.67 | 1015.6 |
|  | Approx. ReLU | 4 | $[-4, 4]$ | 11 | 1.41 | 79.27 | 1463.4 |
|  |  |  | $[-6, 6]$ | 11 | 1.41 | **81.06** | 1555.5 |
|  |  | 2 | $[-4, 4]$ | 8 | 1.15 | 79.06 | 1006.5 |
|  |  |  | $[-6, 6]$ | 8 | 1.15 | 74.79 | 990.2 |

VI. CONCLUSION

In this paper, we proposed a method for improving classification accuracy for CNN inference using HE by adopting BN with a polynomial approximated activation function instead of the square function as an activation function. Moreover, we apply the layer combination and coefficient manipulation of polynomial approximated activation function to reduce the multiplicative depth.

As a result of the evaluation, we achieved a classification accuracy of 99.29% for the MNIST dataset, which outperforms 0.11% compared to the baseline method that uses the square function. Besides, we achieved a classification accuracy of 81.06% for the CIFAR-10 dataset, which outperforms 4.69% compared to the baseline method.

Our future work includes reducing memory usage during inferencing and applying them to deeper network models.

ACKNOWLEDGMENT

This work was supported by JST CREST grant number JPMJCR1503(Japan).